\documentclass[accepted]{article}
\usepackage{microtype}
\usepackage{graphicx}
\usepackage{subfigure}
\usepackage{booktabs} 

\usepackage{hyperref}

\usepackage{icml2022}

\usepackage{amsmath}
\usepackage{amssymb}
\usepackage{mathtools}
\usepackage{amsthm}

\usepackage[capitalize,noabbrev]{cleveref}

\renewcommand{\algorithmicrequire}{\textbf{Input:}}    
\renewcommand{\algorithmicensure}{\textbf{Output:}} 
\newtheorem{theorem}{Theorem}[section]

\newtheorem{coro}[theorem]{Corollary}

\usepackage{booktabs}
\usepackage{multirow}
\usepackage{siunitx}
\usepackage{amssymb}
\usepackage{booktabs}       
\usepackage{amsfonts}       
\usepackage{nicefrac}       
\usepackage{microtype}      
\usepackage{amsmath}
\usepackage{graphicx}
\usepackage{multirow}
\usepackage{booktabs}
\usepackage{wrapfig}

\usepackage[textsize=tiny]{todonotes}

\icmltitlerunning{PACE: A Parallelizable Computation Encoder for Directed Acyclic Graphs}

\begin{document}

\twocolumn[
\icmltitle{PACE: A Parallelizable Computation Encoder for Directed Acyclic Graphs}



\icmlsetsymbol{equal}{*}

\begin{icmlauthorlist}
\icmlauthor{Zehao Dong}{equal,yyy1}
\icmlauthor{Muhan Zhang}{equal,yyy2,yyy3}
\icmlauthor{Fuhai Li}{yyy4}
\icmlauthor{Yixin Chen}{yyy1}
\end{icmlauthorlist}

\icmlaffiliation{yyy1}{Department of Computer Science $ \&$ Engineering, Washington University in St. Louis, St. Louis, USA}
\icmlaffiliation{yyy2}{Institute for Artificial Intelligence, Peking University, Beijing, China}
\icmlaffiliation{yyy3}{Beijing Institute for General Artificial Intelligence, Beijing, China}
\icmlaffiliation{yyy4}{Institute for Informatics and Department of Pediatrics, Washington University in St. Louis, St. Louis, USA}


\icmlcorrespondingauthor{Yixin Chen}{chen@cse.wustl.edu}

\icmlkeywords{Machine Learning, ICML}

\vskip 0.3in
]



\printAffiliationsAndNotice{\icmlEqualContribution} 

\begin{abstract}
   Optimization of directed acyclic graph (DAG) structures has many applications, such as neural architecture search (NAS) and probabilistic graphical model learning. Encoding DAGs into real vectors is a dominant component in most neural-network-based DAG optimization frameworks.
   Currently, most DAG encoders use an asynchronous message passing scheme which sequentially processes nodes according to the dependency between nodes in a DAG. That is, a node must not be processed until all its predecessors are processed. As a result, they are inherently not parallelizable. In this work, we propose a Parallelizable Attention-based Computation structure Encoder (PACE) that processes nodes simultaneously and encodes DAGs in parallel. We demonstrate the superiority of PACE through  encoder-dependent optimization subroutines that search the optimal DAG structure based on the learned DAG embeddings. Experiments show that PACE not only improves the effectiveness over previous sequential DAG encoders with a significantly boosted training and inference speed, but also generates smooth latent (DAG encoding) spaces that are beneficial to downstream optimization subroutines. Our source code is available at \url{https://github.com/zehao-dong/PACE}.
  
\end{abstract}
\section{Introduction}
Directed acyclic graphs (DAGs) are ubiquitous in various real-world problems including neural architecture search~\citep{elsken2019neural,wen2020neural}, source code modeling~\citep{allamanis2018survey}, structure learning of Bayesian networks~\citep{koller2009probabilistic,zhang2019d}, etc. One key challenge in DAG optimization problems is the difficulty to use gradient strategies to quickly adjust the structure of a DAG towards the right direction due to the absence of gradient information. Some earlier works propose to directly optimize the discrete DAG structure through black-box optimization techniques such as reinforcement learning~\citep{zoph2016neural}, evolutionary algorithms~\citep{liu2017hierarchical}, and Bayesian optimization~\citep{kandasamy2018neural}, which are inherently less efficient. A more recent approach is to encode DAGs into some continuous space for searching, and various DAG encoders have been developed. In general, these DAG encoding schemes fall into two categories: structure-aware encoding scheme~\citep{ying2019bench, wen2020neural, shi2020bridging} and computation-aware (performance-aware) encoding scheme~\citep{zhang2019d, Thost2021DirectedAG}. 

Due to the superior graph representation learning ability, graph neural networks (GNNs) have broadly achieved state-of-art performance on various graph learning tasks, such as node classification~\citep{Velickovic2018GraphAN,Hamilton2017},
graph classification~\citep{xu2018how,zhang2018end,duvenaud2015convolutional}, link prediction~\citep{zhang2018link}, and hyperlink prediction~\citep{zhang2018beyond}. Basically, GNNs follow the message passing scheme~\citep{gilmer2017neural} where each node aggregates node features from its one-hop neighborhood repeatedly to update its own feature, and this aggregation happens at all nodes simultaneously. However, \citet{Thost2021DirectedAG} suggests that such a framework cannot exploit the inductive bias of the computation dependency defined by DAGs, thus failing to generate a smooth encoding space beneficial to  downstream optimization and prediction routines.

Hence, in order to model the dependency between nodes in DAGs, various GNNs specifically designed for encoding DAGs, such as D-VAE~\citep{zhang2019d} and DAGNN~\citep{Thost2021DirectedAG}, are developed to inject the computation dependency between nodes into the representation learning process. Instead of updating node features simultaneously, these DAG encoders are constructed upon a gated recurrent unit (GRU) and will not update the representation of a node until all of its predecessors are updated. Such an asynchronous message passing scheme actually simulates how a real computation is performed along the DAG---the message passing order respects the computation dependency defined by the DAG, thus better exploiting the inductive bias. One way to achieve this is to perform message passing sequentially following a topological ordering of the nodes. However, one key limitation of such DAG encoders is that the encoding process is inherently sequential, precluding processing all nodes in parallel. Although DAGNN proposes a topological batching trick to accelerate the training speed by partitioning nodes into disjoint batches where nodes within a batch can be processed in parallel, the time complexity is still lower-bounded by the longest path (diameter) of the DAG and the fundamental constraint of the sequential computation nature still remains. 

Numerous efforts have been made to reduce the sequential computation cost in the sequence modeling literature~\citep{cho2014learning,wu2016google}. For instance, ConvS2S~\citep{gehring2017convolutional} utilizes convolutional layers as building blocks to compute output representations at different positions, while Transformer~\citep{vaswani2017attention} proposes to inject the position (order) information into the model through positional encoding, and then the dependency between positions can be captured through the attention mechanism~\citep{bahdanau2014neural,gehring2017convolutional} in parallel instead of resorting to recurrent neural networks. However, the success of these techniques relies on the inherent linear order of symbols in the input/output sequences that automatically characterizes the dependency between symbols. That is, the dependency between symbols is fully captured by their positions in the sequence. Such a condition is not satisfied by nodes in a DAG since each node can have \textbf{multiple parents} instead of only one like in plain sequences, and the dependency between nodes forms a (strong) partial order rather than a linear order. Thus, previous parallel sequence modeling methods would fail in the representation learning of DAGs. 

In this paper, we propose a novel Parallelizable Computation Encoder, PACE, to improve the efficiency over existing GRU-based DAG encoders. In order to borrow the power of Transformer for sequence modeling to DAG modeling problems, we need to design a \textit{positional encoding} scheme specifically for DAGs which can fully capture the dependency between nodes in a DAG before applying the pairwise self-attention mechanism. To achieve this, we propose a GNN-based dag2seq framework which is proved to injectively map DAGs to sequences of node embeddings. This means, we are able to fully recover the DAG structure from these produced node embeddings, the same as the positional encoding in the original Transformer.
After that, a Transformer encoder (with mask operation) is applied to the node embedding sequence to simultaneously learn representations for
all nodes in the DAG through the self-attention mechanism. This way, PACE incorporates the relational inductive bias~\citep{battaglia2018relational,xu2020can} carried by DAG structures into the encoding process, while eschewing the recurrence in previous works thus greatly improving the parallelization and encoding efficiency.

To demonstrate the superiority of the proposed PACE model, we 
evaluate PACE against current state-of-art DAG encoders, GNN-based graph encoders, and recent (undirected) graph Transformers. Massive experiments show that PACE not only outperforms competitive baselines 
but also significantly boosts the training and inference speed through parallelization.

\section{Backgrounds}
\label{back}
\subsection{Parallelizable Sequence Models}
Encoding the complexities and nuances of sequences plays a central role in various machine learning tasks, including sentiment classification~\citep{medhat2014sentiment}, speech recognition~\citep{abdel2014convolutional}, and other natural language processing (NLP) tasks \citep{khan2016survey}. For many years, sequential models, such as recurrent neural networks (RNNs)~\citep{medsker2001recurrent}, were the primal way to solve the sequence encoding problem. These models are computationally expansive due to the sequential encoding process. Hence, many parallelizable sequence models are proposed, including Transformer~\citep{vaswani2017attention}, BERT~\citep{devlin2019bert}, etc. Our proposed PACE model is built upon the Transformer (encoder) architecture.

Transformer~\citep{vaswani2017attention} is arguably the earliest translation model that solves the sequence-to-sequence task without using sequence-aligned RNNs or convolutional architectures. Transformer relates information from different positions in the sequence through the (masked) self-attention mechanism to encode/decode the representation of items in the input/output sequence, and incorporates the order of the items into the encoding/decoding process through a positional encoding mechanism. Briefly, the positional encoding mechanism is an injective function $f_{pe}: \mathbb{N} \to \mathbb{R}^{d}$ that represents the positions (i.e. indices) of items in the sequence as $d$-dimensional vectors. Hence, it consistently outputs a unique encoding for each position in the sequence. The Transformer architecture is inherently parallelizable and can capture the long-term dependency with ease, thus is broadly applied to sequence modeling tasks in following works~\citep{dai2019transformer,al2019character,devlin2019bert,lewis2020bart}. 

\subsection{DAG Encoding Problem}

\begin{figure*}[t]
    \centering
    \includegraphics[width=0.99\linewidth]{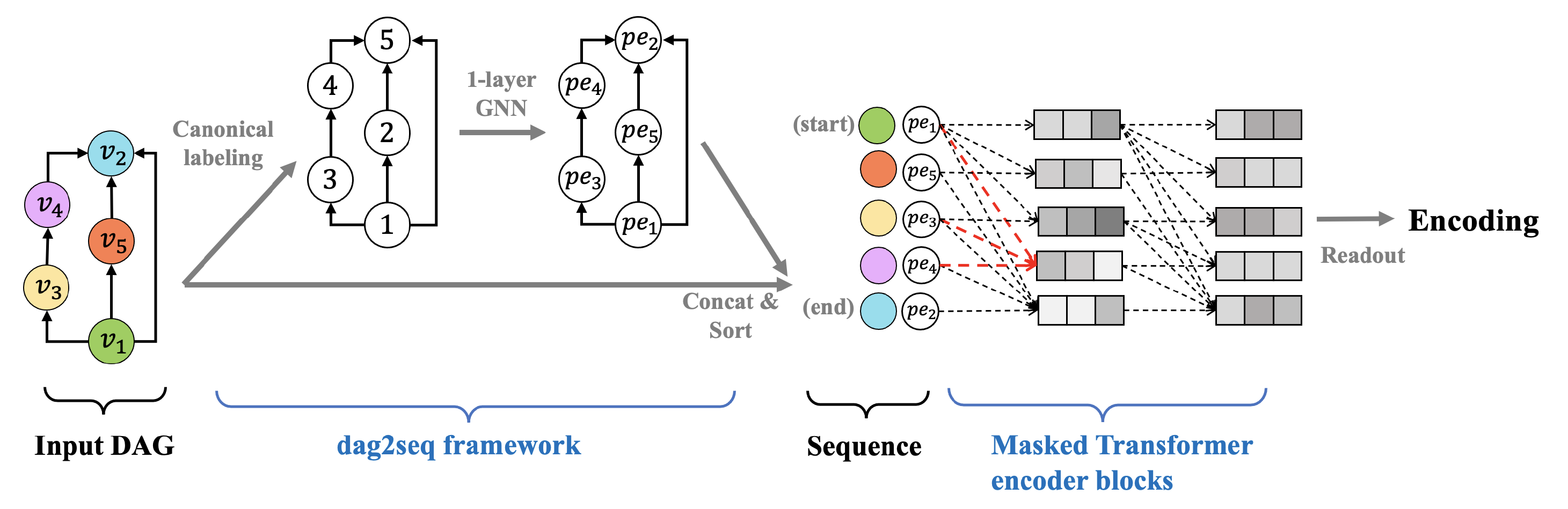}
    \vspace{-10pt}
    \caption{Illustration of PACE. The input DAG is first injectively represented as a sequence through the dag2seq framework, and then the sequence is fed into multiple stacked masked Transformer encoder blocks. The operations of nodes (i.e. node types) are visualized as colors, and nodes in the sequence is sorted according to the canonical label generated in the dag2seq framework.}
    \label{fig:over_arcg}
\end{figure*}

A directed acyclic graph (DAG) $G$ is represented as a pair $(V,E)$ with $V=\{v_{1},v_{2},..,v_{n}\}$ denoting the set of nodes and $E \in V \times V$ denoting the set of directed edges. A DAG often carries a computation. We use $\mathcal{O}$ to denote an (computational) operation dictionary. For instance, the operation dictionary $\mathcal{O}$ for  NAS-Bench-101 dataset contains five operations: ``Input'', ``Output'', ``$3 \times 3$ convolution'', ``$1 \times 1$ convolution'', and ``$3 \times 3$ max-pool''.
Let $G = (V,E)$ be a DAG whose nodes represent operations in $\mathcal{O}$. Then the DAG $G$ represents a \textbf{computation structure} in which  dependencies between operations are determined by directed edges in $E$. Hence, isomorphic DAGs define the same computation structure. Then the objective of the DAG encoding problem is to develop encoders that can generate embeddings to distinguish the computation structures defined by DAGs in the encoding space.

\begin{wrapfigure}[12]{R}{0.55\columnwidth}
  \centering
  \vspace{-14pt}
  \hspace{-18pt}\includegraphics[width=0.56\columnwidth]{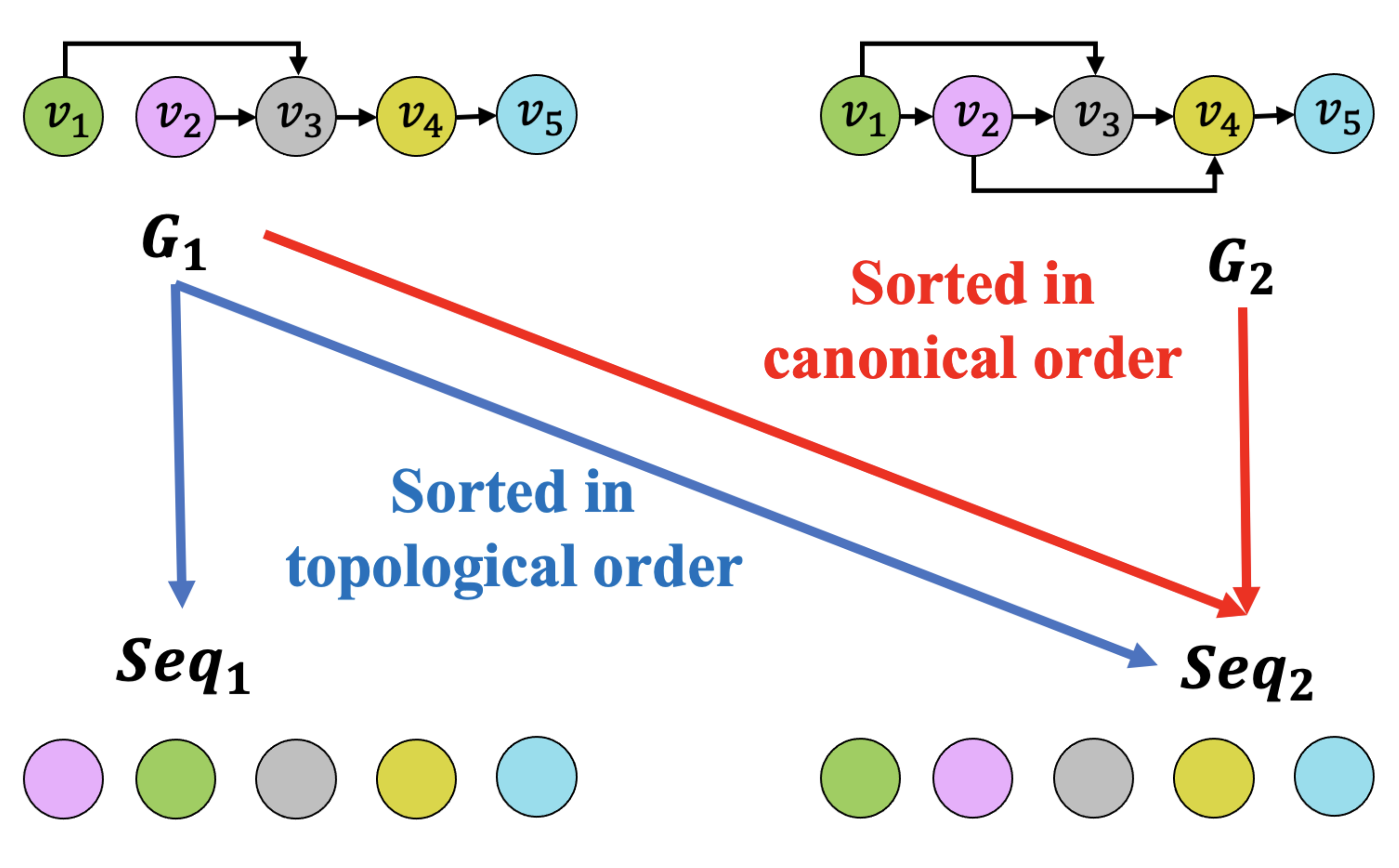}
  \vspace{-10pt}
  \hspace{-17pt}\caption{Illustration of the ambiguity when applying sequence models (such as Transformer) to the DAG encoding problem.}
  \label{fig:illu}
\end{wrapfigure}

DAGs $G=(V,E)$ show a close relation with partial order. For any two nodes $v_{i}, v_{j} \in V$, let $ \preceq $ be a binary relation such that $v_{i} \preceq v_{j}$ if and only if there is a directed path from $v_{i}$ to $v_{j}$, then the binary relation $ \preceq $ defines a partial order on the node set $V$. Based on the partial order, sequential DAG encoders, such as D-VAE~\citep{zhang2019d} and DAGNN~\citep{Thost2021DirectedAG}, use GRU~\citep{cho2014learning} to recursively encode nodes in the input DAG, where a node is not encoded until all of its predecessors (those nodes with a $\preceq$ relation with it) are encoded. Because of the possibly very long dependency chains, these sequential DAG encoders inherently share the same limitations as RNNs, such as the slow training and inference speed and the difficulty to capture long-term dependencies.

To address these limitations, it is intuitive to generalize Transformer to the DAG encoding problem, as Transformer brings undoubtedly a huge improvement over the RNN-based sequence models. However, Figure~\ref{fig:illu} illustrates that it can be ambiguous how to represent DAGs as sequences due to the complex topological structure of DAGs.
Let $o: V \to \mathcal{O}$ be the function that maps  each node in $G$ to an operation in $\mathcal{O}$, and $f_{pe}$ be the original positional encoding function in Transformer. 
One intuitive way to linearize a DAG into a sequence is to sort its nodes with a topological order, which is also used in GRU-based DAG encoders. However, the topological orders of nodes in a DAG are often not unique. For instance, graph $G_{1}$ in Figure~\ref{fig:illu} has two different topological orders: $v_{1}, v_{2}, v_{3}, v_{4}, v_{5}$ and $v_{2}, v_{1}, v_{3}, v_{4}, v_{5}$, hence resulting in two different node sequences $\textit{Seq}_{1}$ and $\textit{Seq}_{2}$. Note that this does not hurt existing GRU-based DAG encoders as they are invariant to which topological order is used.

To avoid the ambiguity of topological order, another approach is to sort nodes according to a canonical order (i.e., node index in the canonical form of the DAG) as suggested by \cite{niepert2016learning}. For example, let $v_1, v_2, v_3, v_4, v_5$ be the nodes sorted by a canonical order. Then, we can linearize a DAG into a sequence $(o(v_{1}),f_{pe}(v_{1})), (o(v_{2}),f_{pe}(v_{2})), ... (o(v_{n}),f_{pe}(v_{n}))$, similar to how Transformer represents a sentence. Although canonical order guarantees the generated sequence is unique for the same DAG, different DAGs might have the same sequence. For instance, DAGs $G_{1}$ and $G_{2}$ in Figure~\ref{fig:illu} are not isomorphic, yet they will be represented as the same sequence $\textit{Seq}_{2}$. This is because the positional encoding function $f_{pe}$ only captures node positions in the sequence but fails to encode the node dependencies, causing a significant structural information loss.

In summary, it is not straightforward to generalize Transformer to DAGs. We need to design a smart linearization method for DAGs which guarantees that the same DAG is transformed to the same sequence, while different DAGs are always transformed to different sequences, in order to achieve a lossless transformation.

\section{The PACE Model}
In this section, we describe the proposed \textbf{Pa}rallelizable \textbf{C}omputation  \textbf{E}ncoder (\textbf{PACE}). The key component of PACE is a \textbf{dag2seq} framework, which leverages graph canonization procedure and a graph neural network to transform DAGs into sequences while preserving the distinctiveness of non-isomorphic DAGs, so that the DAG encoding problem can be addressed efficiently by applying Transformer to the sequences.


\subsection{The Dag2seq Framework}
\label{subsec:dag2seq}

Here we describe the proposed dag2seq framework. Let $G = (V,E,o)$ be a labeled DAG (node labels are operations in $\mathcal{O}$), where $V = \{1,2, ...,n\}$ is the finite set of nodes, $E$ is the set of directed edges, and $o: V \to \mathcal{O}$ is a function that associates to each node an operation in $\mathcal{O}$. We denote the \textit{canonical form} of $G$ as $C(G) = (V^{C}, E^{C}, o^{C})$, which assigns to each labeled graph $G$ an isomorphic labeled graph $C(G)$ that is a unique representation of its isomorphism class. That is, all labeled graphs isomorphic to $G$ will have the same canonical form $C(G)$. Since $G$ and $C(G)$ are isomorphic, there exists a bijection $\pi : V \to V^{C}$ (note that $V^{C}=\{1,2, ...,n\}$) between the node sets such that $o^{C}(\pi(i)) = o(i)$ for all $i \in V$ and $(\pi(i),\pi(j)) \in E^{C}$ if and only if $(i,j) \in E$. The graph canonization process assigns a new index $\pi(i)$ to each node $i$.
Based on the new indices $\pi(i)$, the proposed dag2seq computes the \textit{positional encoding} of node $j \in V$ (denoted as $p_{j}$) as follows
\begin{align}
    a_{j} = \textit{Agg} (\{ \pi(i), (i,j) \in E \}) \label{func:1} \\
    p_{j} = \textit{Combine} (\pi(j), a_{j}) \label{func:2}
\end{align}
where functions \textit{Agg} and \textit{Combine} follow the same definition as Graph Isomorphism Network (GIN)~\citep{xu2018how}. Note that $\pi(i), i \in V$ are discrete variables, and function (\ref{func:1}) and (\ref{func:2}) take their one-hot encodings as input in practice. From Equations (\ref{func:1}) and (\ref{func:2}), we can see that dag2seq uses the canonical indices as node features and applies a one-layer injective GNN to obtain the positional encoding of each node.
Note that $(i,j) \in E$ is equivalent to $(\pi(i),\pi(j)) \in E^{C}$. Hence, Equations (\ref{func:1}) and (\ref{func:2}) can also be interpreted as applying a one-layer injective GNN on the canonical graph $C(G)$ with the true node indices of $C(G)$ as node features. For notation convenience, we use $\pi^{-1}: V^{C} \to V$ to denote the inverse function of $\pi$. Then Theorem~\ref{thm:3.1} describes how dag2seq generates sequences that uniquely represent DAGs.

\begin{theorem}
\label{thm:3.1}
Let $G = (V,E,o)$ be a labeled DAG, and $p_{1}, p_{2}, ..., p_{n}$ be the  positional encodings generated by dag2seq. If functions \textit{Agg} and \textit{Combine} are injective, then the sequence $(o(\pi^{-1}(1))$, $p_{\pi^{-1}(1)})$, $(o(\pi^{-1}(2))$, $p_{\pi^{-1}(2)})$, ..., $(o(\pi^{-1}(n))$, $p_{\pi^{-1}(n)})$ injectively encodes the computation structures defined by DAGs.  
\end{theorem}

We prove Theorem \ref{thm:3.1} in Appendix \ref{sec:thm:3.1}. Theorem \ref{thm:3.1} provides a guarantee that node sequences generated by dag2seq injectively encode the original DAGs which fully preserve the node type information as well as structure information of the original DAGs. In other words, two labeled DAGs will be encoded into the same sequence if and only if they are isomorphic (i.e. their computation structures are the same). Then advanced parallelizable encoders (such as the Transformer encoder) for sequence modeling can be applied to the DAG encoding problem to improve the efficiency of the encoding process, thus facilitating the downstream optimization and prediction routines.
Intuitively, the canonical form provides a unified node indexing for isomorphic DAGs which initially may have different node orderings, while the one-layer injective GNN encodes the direct predecessors of each node into its positional encoding. Then it is not difficult to see that from the canonical indices of nodes and their direct predecessors as well as the node types we can fully recover the original DAG. The one-layer GNN is parallelizable, in contrast to the sequential GNNs used in previous works~\citep{zhang2019d,Thost2021DirectedAG}.

It is also worth discussing the complexity of graph canonization. The graph canonization problem is theoretically at least as computationally hard as the graph isomorphism problem, which is in NP but not known to be solvable in polynomial time nor to be NP-complete. However, getting the canonical form of graphs is not too difficult in practice, thanks to the well-known graph canonization tools such as Nauty~\citep{mckay2014practical}. Empirically, such tools usually return the canonical form of a reasonable-sized graph in seconds. Theoretically, Nauty has an average time complexity of $O(n)$, and polynomial-time graph canonization algorithms also exist for graphs of bounded degrees. We also found in our experiments that graph canonization adds a negligible overhead.

\subsection{The Transformer in PACE}

With the dag2seq framework to injectively map DAGs to sequences, we next develop the attention-based parallelizable encoder which consists of $K$ stacked Transformer (encoder) blocks.

For each item $(o(\pi^{-1}(i)), p_{\pi^{-1}(i)})$ in the output sequence of dag2seq, $ o(\pi^{-1}(i)) $ provides the operation information, while the positional encoding $p_{\pi^{-1}(i)}$ contains the structural information. Hence, PACE concatenates the trainable embedding of operation $ o(\pi^{-1}(i)) $ and the positional encoding $p_{\pi^{-1}(i)}$ as the embedding $e_i$ of item $i$ in the sequence.
\begin{align}
    e_i = \textit{Concat}(\textit{Emb}(o(\pi^{-1}(i))), p_{\pi^{-1}(i)})
\end{align}
Then the sequence $e_1, e_2,...,e_n$ is fed into the first Transformer encoder block. Each transformer encoder block performs the multi-head self-attention mechanism~\citep{vaswani2017attention} to update the embedding of each item in the sequence. We provide details of the multi-head self-attention mechanism in Appendix~\ref{sec:multi_head}.

In the original transformer encoder blocks, the attention operations are not masked. In other words, for any two items $i$ and $j$ in the sequence, the embedding of item $i$ will be used to update the embedding of item $j$. Such unmasked operation is reasonable as the positional encodings $\{p_{\pi^{-1}(i)}, \forall i \}$ have already encoded the partial order between nodes. However, due to the complex dependencies a DAG may encode, fully relying on the positional encodings to capture such dependencies might not be enough. Therefore, we introduce a masked attention operation that helps better capture the dependency between nodes in practice. The masked attention operation can be specified through a binary mask matrix $M$ such that $M_{i,j} = \textit{False}$ if there exists a path from $i$ to $j$ in $C(G)$ and $M_{i,j} = \textit{True}$ otherwise.

In this mask matrix $M$, element ${M}_{i,j} = \textit{True}$ indicates that the effect of item $i$ in updating the embedding of item $j$ will be masked out. When ${M}_{i,j} = \textit{True}$, there is no (directed) path from node $i$ to node $j$ in the canonical form $C(G)$. 
In other words, we only allow $j$'s predecessors in $C(G)$ (or equivalently, $\pi^{-1}(j)$'s predecessors in $G$) to participate in the updating of $j$'s embedding. Such a masking operation has two benefits: 1) the structure information of the DAG is strengthened in the masked self-attention, and 2) the partial order between operations in the computation structure is exploited, which aligns with the logic of a real computation in the sense that the operation at some node does not depend on its successor operations. We also empirically verify the effectiveness of the masked attention operation in the ablation study.
The mask matrix $M$ can be efficienly computed through the DFS algorithm or the Floyd algorithm, which we describe in Appendix \ref{sec:flo}. 

Together with the proposed dag2seq framework, Transformer in PACE helps reduce the complexity of covering the entire DAG's diameter significantly.
The positional encoding (from dag2seq) of each node encodes its canonical index and its incoming neighbors. Since the masked self-attention aggregates messages from all predecessors of a node, this node will know all its predecessors and their connection structures in one step, which provides us all the information needed to compute graph metrics such as shortest path distance to this node with only one message passing step. In contrast, ordinary GNNs require at least $O(\text{shortest path distance})$ message passing steps to capture the dependency/attention between two nodes. For the last node to capture the entire graph structure, GNNs will require $O(\text{diameter})$ message passing steps. Therefore, PACE allows using much shallower layers than GNNs to encode DAGs (especially when the DAG has a long diameter) and increases the speed significantly.


\subsection{Training Methodology}
We design two schemes to train the PACE model. One is to train a variational autoencoder (VAE) for DAGs that is able to encode and decode/generate DAGs into and from a latent space, like D-VAE.
On the other hand, since the proposed dag2seq transforms the DAG encoding problem to the sequence encoding problem, pre-training (self-supervised learning) techniques in NLP are also suitable for training PACE.

Without loss of generality, we assume that there is a single output node in each DAG that has no successor. If not, we can add a virtual output node and add directed edges from all nodes whose out-degree is $0$ to the output node. When PACE is trained in a VAE architecture, we use a common trick in standard Transformers by assuming that there are at most $N$ nodes in the input DAG. If a DAG has $n < N$ nodes, we pad $N-n$ end symbols to the end of the sequence generated by dag2seq. Then PACE readouts the DAG encoding by concatenating the learned embeddings of the $N$ symbols. When PACE is trained in a pre-training architecture, similar to the sentiment classification task in BERT, PACE takes the learned embedding of the output node as the DAG encoding.

\noindent\textbf{Training PACE in a VAE architecture.}~~In the PACE-VAE architecture, we take PACE as the encoder, and connect the output of PACE with two fully connected (FC) layers to predict the mean and variance of the approximated posterior distribution in the evidence lower bound (ELBO)~\citep{kingma2013auto}. The decoder of PACE-VAE consists of $K=3$ standard Transformer decoder blocks of dimension $d_{k}$. Given the latent vector $z$ to decode, the decoder uses a FC layer to reconstruct a vector of dimension $N \times d_{k}$. The vector is then reshaped to a matrix $Z$ of shape $(N, d_{k})$, which plays the same role as the ``memory'' matrix in a standard Transformer in NLP. Hence, the encoder-decoder attention layer in each Transformer decoder block takes the decoded matrix $Z$ as the ``Key'' matrix and ``Value'' matrix, and uses the output from the previous self-attention layer as the ``Query'' matrix. 
Similar to the standard Transformer, during the generation, the decoder of PACE-VAE sequentially predicts nodes in $G$ according to the learnt canonical order, and this process is ended until a special symbol is predicted indicating the decoding process is completed. For each generated node $i$ in the decoding process, we do a softmax to select the operation of the node, and use a binary classifier to predict the existence of an edge between node $i$ and any node $j < i$. We describe details about PACE-VAE and its parallelizable training framework in Appendix~\ref{sec:overall_detail}.

\noindent\textbf{Training PACE in a pre-training architecture.}~~After converting DAGs to sequences by the proposed dag2seq framework, PACE is essentially a sequence modeling encoder. \cite{yan2020does} validates that the pre-training architecture in NLP that generates embeddings without using accuracies can better preserve the local structural relationship in the latent space. As such, PACE can also take the masked language modeling (MLM)~\citep{devlin2019bert,yan2021cate} objective for (pre-)training to capture the locality information of the computation structure defined by DAGs. For each input DAG, we randomly select 20\% nodes for masking and prediction, where 80\% of them are replaced with the $[\textit{MASK}]$ token and the remaining nodes are unchanged. The output embeddings are used to predict the original node operations $o(\pi^{-1}(i))$, and we train PACE by minimizing the cross-entropy loss of the predicted node operations and the true node operations.

\section{Comparison to Related Works}
Despite the great success of Transformers in modeling sequential data such as natural languages and images \cite{carion2020end,dosovitskiy2020image}, there has been no work generalizing Transformers to another important type of sequential data, namely DAGs. Existing DAG encoders mostly adopt an RNN-style encoding scheme to sequentially process nodes in a DAG. S-VAE~\citep{bowman2016generating} takes as input the sequence of node strings which consist of the node type as well as the adjacency vector of each node, and then applies a GRU-based RNN to the topologically sorted node sequence to encode a DAG. To encode the semantics of the computation carried by a DAG, D-VAE~\citep{zhang2019d} proposes an DAG-style message passing framework that sequentially updates the DAG encoding following a topological order of nodes. For each node in a DAG, D-VAE takes a gated summation aggregator to combine information from all its direct predecessors, and then a GRU is used to update the node embedding based on the aggregated information and the node type. Similar to the encoder of D-VAE, \citet{Thost2021DirectedAG} proposes DAGNN which also sequentially aggregates node features. It differs from D-VAE in the sense that the aggregator is constructed through the attention mechanism and it comes with a layer notion in the DAG encoding process. However, the sequential nature of these RNN-style DAG encoders precludes their parallelizability. In contrast, PACE adopts a dag2seq framework that encodes the dependencies between nodes in the positional encodings, and then applies a Transformer to the node sequence to encode a DAG parallelly.

Various previous works (e.g., S-VAE and GraphRNN) have proposed to linearize graphs to node sequences by adopting different node ordering techniques. S-VAE topologically sorts nodes, where each node feature is the concatenation of the one-hot encoding of the node type and a 0/1 vector indicating whether (directed) edges exist from previous nodes to itself. On the other hand, GraphRNN takes the BFS ordering and utilizes the fixed $M$-dimensional vector to represent the node connectivity in the BFS queue. The above methods share a major limitation---both topological ordering and BFS ordering are not unique. That is, the same DAG can still be represented as different sequences, thus generating different DAG embeddings. Consequently, similar DAGs might be encoded to vastly different vectors in the DAG encoding space, which makes the encoding space unsmooth and the DAG optimization problem difficult. In contrast, our dag2seq framework injectively represents DAGs as sequences, which guarantees that the same DAG is always represented as the same sequence, and different DAGs are always represented as different sequences. This is crucial for the downstream DAG optimization problems.

Concurrent to our work, a large body of works are proposed to apply Transformer to general graph data. GT \citep{dwivedi2020generalization} and SAN \citep{kreuzer2021rethinking} propose to use the Laplacian positional encodings (Laplacian PEs) \citep{dwivedi2020generalization} to encode the relative positions of nodes in graphs, instead of the original positional encoding framework for sequences. On the other hand, GraphiT \citep{mialon2021graphit} and Graphormer \citep{ying2021transformers} incorporate the (relative) position of nodes in graphs in the attention mechanism. The difference is that GraphiT uses graph kernels to adjust the attention scores, while Graphormer utilizes the shortest path between nodes to do so. Furthermore, \cite{jain2021representing} applies a Transformer module after a standard GNN module to capture the both local and long-term context.
However, all previous graph Transformers have problems when applying to DAGs. Laplacian PEs are constructed from the $k$ smallest non-trivial eigenvectors of the graph Laplacian, which is inherently not suitable for DAGs because graph Laplacian is only defined for undirected graphs. Other relative-position-based approaches such as graph kernels and shortest path are symmetric in nature and do not take the absolute (canonical) index of each node, thus are not injective. In contrast, our positional encoding framework, dag2seq, is DAG-friendly, and injectively converts DAGs into sequences, which preserves the full structure and type information of DAGs so that we can safely treat DAG encoding as sequence encoding. 

\section{Experiments}
\label{exper}

\begin{table*}[t]
\caption{Predictive performance on NA and BN.}
\label{tab:pred}
\centering
\Large
\resizebox{0.7\textwidth}{!}{
\begin{tabular}{lccccc}
\toprule
& \multicolumn{2}{c}{NA} & & \multicolumn{2}{c}{BN} \\
\cmidrule(r){2-3}  \cmidrule(r){5-6} 
Evaluation Metric  & RMSE $\downarrow$ & Pearson's r $\uparrow$ & & RMSE $\downarrow$ & Pearson's r $\uparrow$\\
\midrule
PACE (our model) & \textbf{0.254} $\pm$ \textbf{0.002} & \textbf{0.964} $\pm$ \textbf{0.001 } & & \textbf{0.115} $\pm$ \textbf{0.004} & \textbf{0.994} $\pm$ \textbf{0.001} \\
\midrule
DAGNN & 0.264 $\pm$ 0.004 & 0.964 $\pm$ 0.001 & &  0.122  $\pm$ 0.004 & 0.991 $\pm$ 0.000  \\
D-VAE & 0.384 $\pm$ 0.002 & 0.920 $\pm$ 0.001 & & 0.281 $\pm$ 0.004 & 0.964 $\pm$ 0.001  \\
S-VAE & 0.478 $\pm$ 0.002 & 0.873 $\pm$ 0.001 & & 0.499 $\pm$ 0.006 & 0.873 $\pm$ 0.002  \\
GraphRNN & 0.726 $\pm$ 0.002  & 0.669 $\pm$ 0.001 & & 0.779 $\pm$ 0.007 & 0.634 $\pm$ 0.001  \\
DeepGMG  & 0.478 $\pm$ 0.002  & 0.873$\pm$ 0.001 & & 0.843 $\pm$ 0.007 & 0.555 $\pm$ 0.003  \\
GCN & 0.832 $\pm$ 0.001 & 0.527 $\pm$ 0.001 & & 0.599 $\pm$ 0.006 & 0.809 $\pm$ 0.002  \\
gated-GCN& 0.416 $\pm$ 0.002 & 0.891 $\pm$ 0.001 & & 0.311$\pm$ 0.003 & 0.953 $\pm$ 0.002 \\
\midrule
GT & 0.329$\pm$ 0.001 & 0.942 $\pm$ 0.001 & &0.166 $\pm$ 0.003 & 0.987 $\pm$ 0.000 \\
SAN & 0.311$\pm$ 0.003 & 0.950 $\pm$ 0.001 & &0.158$\pm$ 0.005 & 0.989 $\pm$ 0.001  \\
Graphormer & 0.352 $\pm$ 0.002 & 0.936 $\pm$ 0.001 & &0.181$\pm$ 0.004 & 0.971 $\pm$ 0.001 \\
GraphiT &  0.299 $\pm$ 0.002 & 0.955 $\pm$ 0.001 & &0.142 $\pm$ 0.005 & 0.990 $\pm$ 0.003  \\
\bottomrule
\end{tabular}
}
\end{table*}

\begin{table*}[t]
\caption{Downstream search performance on NAS101 and NAS301.}
\label{tab:down_stream}
\centering
\Large
\resizebox{0.7\textwidth}{!}{
\begin{tabular}{lccccc}
\toprule
& \multicolumn{2}{c}{NAS101 (Regret)} & & \multicolumn{2}{c}{NAS301 (Acc)} \\
\cmidrule(r){2-3}  \cmidrule(r){5-6} 
Search Method  &  DNGO ($\%$) $\downarrow$ & DNGO-LS ($\%$) $\downarrow$ & & DNGO ($\%$) $\uparrow$ & DNGO-LS ($\%$) $\uparrow$\\
\midrule
PACE (our model) &  \textbf{0.391} $\pm$ \textbf{0.241} & \textbf{0.278} $\pm$ \textbf{0.178}& &\textbf{94.507} $\pm$ \textbf{0.165}  & 94.547 $\pm$ 0.145 \\
\midrule
DAGNN &  0.445 $\pm$ 0.224 & 0.448 $\pm$ 0.127& &  94.445 $\pm$ 0.219 & 94.433 $\pm$ 0.156  \\
D-VAE &  0.439 $\pm$ 0.203 & 0.430 $\pm$ 0.222 & & 94.453 $\pm$ 0.148   & 94.428 $\pm$ 0.131  \\
S-VAE &  0.458 $\pm$ 0.175 & 0.451 $\pm$ 0.225 & &  94.332 $\pm$ 0.183  & 94.371 $\pm$ 0.203 \\
GIN  & 0.593 $\pm$ 0.177  & 0.518 $\pm$ 0.201 & & 94.451 $\pm$ 0.224  & 94.411 $\pm$ 0.198  \\
GAT   &  0.597 $\pm$ 0.269  & 0.509 $\pm$ 0.187 & & 94.430 $\pm$ 0.171   & 94.421 $\pm$ 0.202   \\
GCN  &  0.627 $\pm$ 0.161 & 0.538 $\pm$ 0.233 & & 94.448 $\pm$ 0.149   &  94.404 $\pm$ 0.160 \\
gated-GCN & 0.451 $\pm$ 0.177 & 0.428 $\pm$ 0.168 & & 94.461 $\pm$ 0.170 & 94.421  +- 0.164 \\
\midrule
GT & 0.573$\pm$ 0.276 & 0.460 $\pm$ 0.148 & &94.421 $\pm$ 0.174 & 94.533 $\pm$ 0.139 \\
SAN & \textbf{0.390} $\pm$ \textbf{0.250} & 0.291$\pm$ 0.166 & &94.446$\pm$ 0.191 & 94.501 $\pm$ 0.143  \\
Graphormer & 0.429$\pm$ 0.302 & 0.314 $\pm$ 0.182 & &94.477$\pm$ 0.142 & \textbf{94.551} $\pm$ \textbf{0.137} \\
GraphiT &  0.407$\pm$ 0.233 & 0.307$\pm$ 0.181 & &94.482$\pm$ 0.156 & 94.489 $\pm$ 0.138  \\
\bottomrule
\end{tabular}
}
\end{table*}

In this section, we conduct experiments on popular DAG encoding datasets to validate the effectiveness and efficiency of the proposed PACE model against state-of-art GRU-based DAG encoders, general-purpose GNN-based graph encoders, and recent (undirected) graph Transformers. 

\subsection{Datasets and Metrics}
\label{subsec:datasets}

\textbf{NA and BN.} The dataset NA consists of approximately 19K neural architectures generated by the software \texttt{ENAS} \citep{pham2018efficient}. Each architecture has its pre-computed weight-sharing (WS) accuracy on CIFAR-10 \citep{krizhevsky2009learning} and includes 8 nodes.  
The dataset BN consists of 200K Bayesian networks randomly generated by the \texttt{bnlearn} package \citep{scutari2010learning}. Each Bayesian network  has 8 nodes and is associated with a Bayesian Information Criterion (BIC) score that measures the architecture performance on dataset Asia \citep{lauritzen1988local}. 
Following the experimental settings used in \citep{zhang2019d}, PACE is evaluated under a VAE architecture, and we take $90 \%$ NA/BN data as the training set and hold out the rest for testing. To make a fair comparison, we evaluate the quality of DAG encoders by measuring the predictive performance as well as the downstream search performance. Briefly, a sparse Gaussian process (SGP) regression model \citep{snelson2005sparse} is trained to predict the DAG performance from its encoding, and we use rooted mean square error (RMSE) and Pearson correlation (Pearson'r) as metrics to evaluate the predictive performance. When evaluating the downstream search performance, we perform Bayesian optimization (BO) in the DAG encoding space based on the SGP regression model, and compare the performance of the best searched architecture.

\textbf{NAS101 and NAS301.} NAS101 (NAS-Bench-101) and NAS301 (NAS-Bench-301) are two well-known neural architecture search (NAS) benchmark datasets. NAS101 \citep{ying2019bench} consists of approximately 420K neural architectures with pre-computed validation and test accuracies on CIFAR-10, where each architecture has up to 7 nodes and 9 edges.
NAS301 \citep{siems2020bench} is a surrogate benchmark for DARTS \citep{liu2018darts}. 
Following \cite{liu2018progressive,yan2021cate}, we randomly sample 1M neural architectures, where each architecture contains at most 15 nodes. On NAS101 and NAS301, the PACE model is trained with the pre-training architecture using the MLM objective. Since better predictive performance of DAG encoders always facilitates the downstream search, we implement two popular BO-based downstream search methods, DNGO \citep{snoek2015scalable} and DNGO-LS \citep{yan2021cate}, and compare the downstream search performance of different encoders. For NAS101, following the original work of \cite{ying2019bench}, we take the regret as the evaluation metric, where the regret is the difference between the test accuracy of the (offline) best neural architecture and that of the best searched neural architecture (after 20 rounds). For NAS301, since we do not have an oracle for the (offline) best neural architecture, we use the test accuracy of the best searched neural architecture, instead.

\textbf{OGBG-CODE2.} 
Encoding ASTs (abstract syntax trees) is another application area for DAG-based processing. Hence, we also evaluate PACE on dataset OGBG-CODE2 \cite{hu2020open} against state-of-art GRU-based DAG encoders (i.e. D-VAE and DAGNN). 
Dataset OGBG-CODE2 contains approximately 450K Python functions parsed into DAGs. These DAGs in average contain more than 120 nodes, and the largest DAG has more than 30000 nodes.
In the experiment, PACE takes the learnt representation of the output node as the whole-graph embedding for the token prediction task (TOK), and the test F1 score is used as the evaluation metric.

\subsection{Baselines and Model Configuration}
We benchmark the proposed PACE with 1) recent (undirected) graph Transformers (i.e. GT \citep{dwivedi2020generalization}, SAN \citep{kreuzer2021rethinking}, GraphiT \citep{mialon2021graphit} and Graphormer \citep{ying2021transformers}); 2) DAG encoders (i.e.  GraphRNN \citep{you2018graphrnn}, DeepGMG \citep{li2018learning}, S-VAE, D-VAE and DAGNN); and 3) GNN-based graph encoders (i.e. gated GCN \citep{bresson2017residual}, GCN \citep{kipf2016semi}, GIN \citep{xu2018how}, GAT \citep{Velickovic2018GraphAN}).
In the experiments, PACE uses 3 Transformer encoder blocks to boost the training and inference speed. The dimension of the embedding layer that maps node types to embeddings is $64$. The output dimension of the 1-layer GNN in dag2seq is also $64$. On NA and BN, we concatenate the positional encodings and node type embeddings as the node features fed into the first Transformer encoder block. On NAS101 and NAS301, we use the summation of positional encodings and node type embeddings, instead. All the experiments are done on NVIDIA Tesla P100 12GB GPUs.

\subsection{Experimental Results}






\textbf{NA and BN: Table \ref{tab:pred}.} In the experiment, PACE achieves the smallest RMSE and the largest Pearson's r on both NA and BN, indicating that PACE generates the smoothest encoding space with respect to the computation structure defined by DAGs. In addition, the improvement is more significant on NA. One possible reason is that DAGs in NA always have a Hamiltonian path which introduces the long-term dependencies between nodes. Since the Transformer-style attention in PACE can better capture the long-term dependency than GRU-based DAG encoders, PACE can preserve the similarity of DAGs better in the learnt DAG encoding space. Furthermore, Appendix \ref{sec:visual_detect} and Appendix \ref{sec:recon_and_gene} compare the downstream search performance and generation performance, respectively. The results show that PACE still achieves the best performance, and the observation suggests that PACE has superior DAG-encoding ability than competitive encoders.


\textbf{NAS101 and NAS301: Table \ref{tab:down_stream}.} In the experiment, PACE significantly outperforms other baselines. When applying DNGO-lS search on NAS301, PACE achieves the second best performance. In other cases, PACE has the best downstream search performance. Similar to PACE, DAGNN and GAT also use the attention mechanism to model the dependencies (relations) between nodes. However, the attention mechanism in these encoders is only applied to nodes and their direct predecessors (DAGNN) or adjacent nodes (GAT), hence making it hard to capture the long-term dependencies between nodes. On the contrary, PACE allows all predecessive nodes to participate in the attention mechanism while encoding the neighbors through positional encoding, which enables PACE to learn both the long-term dependencies and short-term dependencies with ease, thus effectively capturing the similarity of computation structure defined by DAGs. In addition, Laplacian PEs is dependent on the factorization of the graph Laplacian matrix, hence techniques based on Laplacian PEs (i.e. GT, SAN, and gated-GCN) are inherently not suitable for DAGs, and their performance are not comparable to the proposed DAG-friendly PACE model. 

\textbf{OGBG-CODE2: Table \ref{tab:ogbg-code}.} The results interestingly reveal that PACE effectively learn the long-term dependencies between nodes in a DAG. In this experiment, the training process always takes a batch size of 16 and a training epoch of 30, thus we use the average training and inference time (per epoch) to measure the encoding speed. 
Experimental results show that PACE significantly improves the predictive performance over the state-of-art GRU-based DAG encoders (i.e. DAGNN and D-VAE), and PACE only requires about $\frac{1}{3}$ training time and $\frac{1}{3}$ inference time. More details about the experiment are provided in Appendix \ref{sec:ogbg-code}.

\begin{table}[h]
\caption{OGBG-CODE2: PACE versus state-of-art DAG encoders}
\vspace{3pt}
\label{tab:ogbg-code}
 \Large
\resizebox{0.49\textwidth}{!}{
\begin{tabular}{lcccc}
\toprule
Methods  & Test F1 score $\uparrow$ & Training time (min) $\downarrow$ & Inference time (min) $\downarrow$ \\
\midrule
PACE & \textbf{0.1779} $\pm$ \textbf{0.0021} & \textbf{40.71} $\pm$ \textbf{0.19} & \textbf{2.11} $\pm$ \textbf{0.22}\\
\midrule
DAGNN & 0.1751 $\pm$ 0.0049& 166.19 $\pm$ 3.52& 6.77 $\pm$ 0.36\\
D-VAE & 0.1596 $\pm$ 0.0041 &  134.54 $\pm$ 3.06 & 6.08 $\pm$  0.16\\
\bottomrule
\end{tabular}
}
\end{table}

\begin{table*}[h]
\Huge
\caption{Ablation study.}
\label{tab:abla}
\centering
\resizebox{\textwidth}{!}{
\begin{tabular}{lccccccccccc}
\toprule
& \multicolumn{2}{c}{NA} & & \multicolumn{2}{c}{BN} & & \multicolumn{2}{c}{NAS101 (Regret)} & & \multicolumn{2}{c}{NAS301 (Acc)}\\
\cmidrule(r){2-3}  \cmidrule(r){5-6} \cmidrule(r){8-9} \cmidrule(r){11-12} 
Model configuration & RMSE $\downarrow$ & Pearson's r $\uparrow$ & & RMSE $\downarrow$ & Pearson's r $\uparrow$ & & DNGO ($\%$) $\downarrow$ & DNGO-LS ($\%$) $\downarrow$ & & DNGO ($\%$) $\uparrow$ & DNGO-LS ($\%$) $\uparrow$\\
\midrule
Model1: dag2seq $\&$ Mask  & \textbf{0.254}  & 0.964 & & \textbf{0.115} & \textbf{0.9942} & & \textbf{0.391} & \textbf{0.278} & & \textbf{94.507} & \textbf{94.547}\\
Model2: dag2seq $\&$ No Mask  & 0.255   & \textbf{0.967}  & & 0.119 & 0.9941 & & 0.479 & 0.368 & & 94.501 & 94.505\\
Model3: No dag2seq $\&$ No mask  & 0.981  & 0.001   & & 0.368 & 0.9318 & & 0.600 & 0.498 & & 94.467 & 94.401\\
\bottomrule
\end{tabular}
}
\vskip -0.1in
\end{table*}

\begin{table*}[h]
\caption{Comparison of different DAG-to-sequence methods + Transformer}
\label{table:abla2}
\centering
\Huge
\resizebox{\textwidth}{!}{
\begin{tabular}{lccccccccccccccc}
\toprule
& \multicolumn{2}{c}{NA} & & \multicolumn{2}{c}{BN} & & \multicolumn{2}{c}{NAS101 (regret)} & & \multicolumn{2}{c}{NAS301 (acc)} & & \multicolumn{2}{c}{ogbg-code2} \\

\cmidrule(r){2-3}  \cmidrule(r){5-6} \cmidrule(r){8-9} \cmidrule(r){11-12} \cmidrule(r){14-15}

& RMSE $\downarrow$ & Pearson's r $\uparrow$ & & RMSE $\downarrow$ & Pearson's r $\uparrow$ & & DNGO ($\%$) $\downarrow$ & DNGO-LS ($\%$) $\downarrow$ & & DNGO ($\%$) $\uparrow$ & DNGO-LS ($\%$) $\uparrow$ & & Test F1 score $\uparrow$\\
\midrule
PACE & \textbf{0.254} & \textbf{0.964} & & \textbf{0.115} & \textbf{0.994}  & & \textbf{0.391} & \textbf{0.278} & & \textbf{94.507} & \textbf{94.547} & & \textbf{0.1779}\\
\midrule
S-VAE (Transformer) & 0.392 & 0.903 & & 0.417 & 0.901 & & 0.439 & 0.386 & & 94.361 & 94.384 & & 0.1481\\
GraphRNN (Transformer) & 0.406 & 0.896 & & 0.431 & 0.889 & & 0.427 & 0.371 & & 94.410 & 94.392 & & 0.1463\\
\bottomrule
\end{tabular}
}
\end{table*}

\subsection{Computational Cost}

\begin{figure}[t]
\begin{center}
\centerline{\includegraphics[width=0.85\columnwidth]{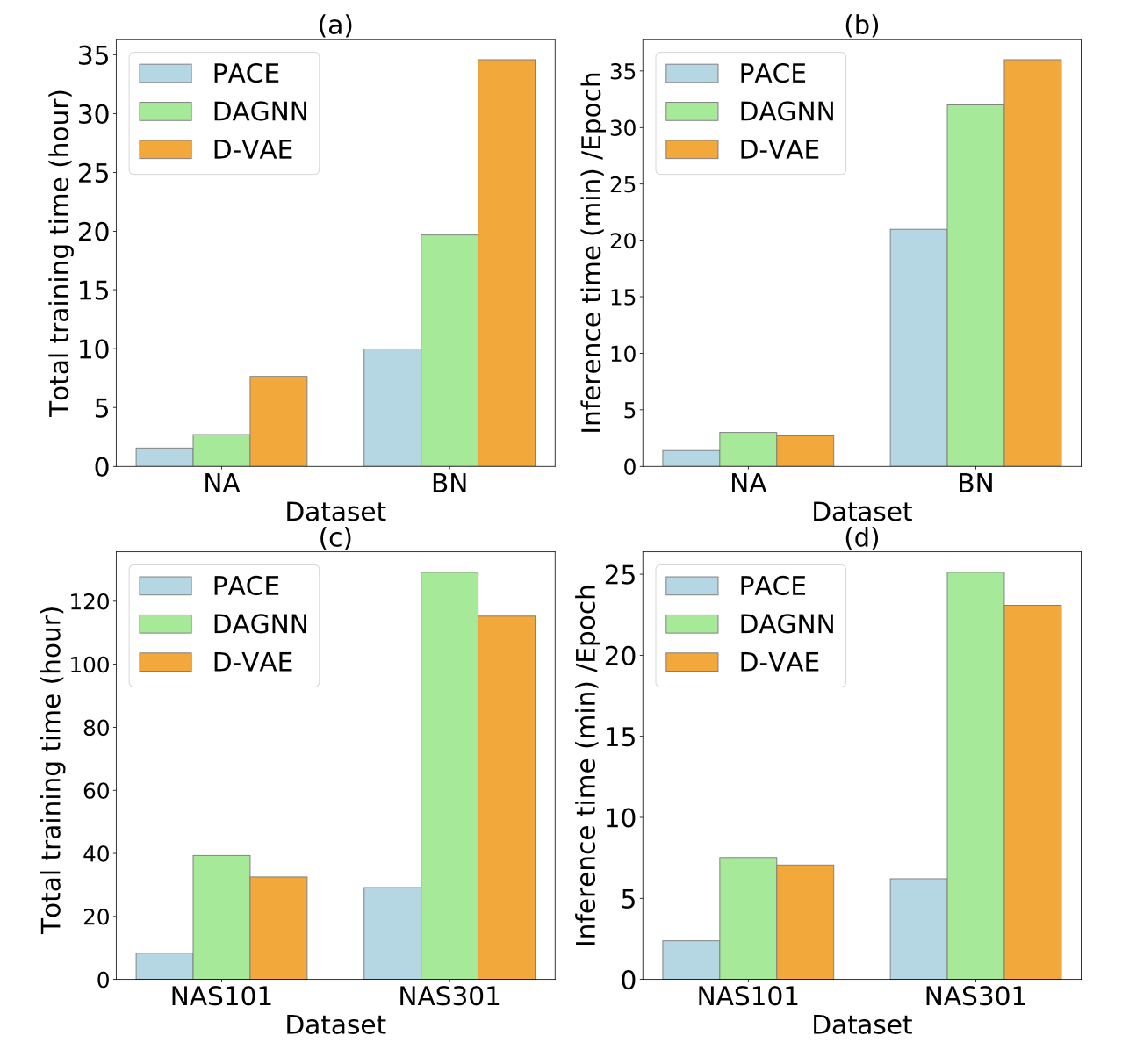}}
\vskip -0.16in
\caption{Computational cost.}
\label{fig:cost_toge}
\end{center}
\vskip -0.36in
\end{figure}

A key advantage of PACE is the parallelizable DAG encoding process, so we compare the computational cost of PACE to GRU-based DAG encoders (D-VAE and DAGNN). We use a single GPU for each experiment, and \textbf{Figure \ref{fig:cost_toge}} shows our results. In this experiment, PACE is trained with the VAE architecture on datasets NA and BN, but with BERT-like objective (the pre-training architecture) on datasets NAS101 and NAS031. On the other hand, D-VAE and DAGNN are always trained with a VAE architecture. Then, to make a fair comparison, we compare the total training time and the average inference time per epoch to evaluate the computational cost of each method. Figure \ref{fig:cost_toge} shows that PACE requires about $\frac{1}{3}$ total training time and about $\frac{1}{3}$ average inference time compared to GRU-based DAG encoders (i.e. D-VAE and DAGNN). Hence, PACE significantly boosts the DAG encoding speed.  


\subsection{Ablation Study}

 In the ablation study, we demonstrate the effectiveness of our proposed dag2seq (positional encoding) framework and the attention mask in PACE. From \textbf{Table \ref{tab:abla}}, we have the following observations: 1) In general, PACE trained with attention mask outperforms the one without attention mask, indicating that the attention mask helps better capture the inductive bias of DAGs. Nevertheless, even without attention mask, PACE still performs relatively well because the dag2seq framework also captures the node dependencies. 2) We also find that the dag2seq framework is vital for the performance of PACE. Without dag2seq, PACE (without mask) almost completely fails at the NA dataset. This verifies the importance of dag2seq for solving the ambiguity issue of DAG encoding illustrated in Figure~\ref{fig:illu}. 
 
 \textbf{Comparison of different DAG-to-sequence methods} To further decouple the contribution of dag2seq from that of the Transformer, we replace the RNN encoders in the baselines S-VAE and GraphRNN with Transformers, and compare them with our PACE model (dag2seq + Transformer).
 GraphRNN is originally designed as a pure generative model, and thus does not have a DAG encoder. However, following the idea of S-VAE, we generate the node feature sequence by replacing the topological ordering in S-VAE with the BFS ordering as in GraphRNN, and replacing the 0/1 vector in S-VAE with the fixed $M$-dimensional vector. \textbf{Table \ref{table:abla2}} shows the results. PACE still demonstrates significant performance advantages, illustrating the importance of dag2seq for DAG encoding. 

\section{Conclusion}
\label{convlu}
In this paper, we have proposed PACE, a novel DAG encoder based on Transformer. 
Unlike traditional GRU-based DAG encoders which sequentially encode DAG nodes, PACE is fully parallelizable and effectively learn the long-range dependencies of node pairs in a DAG. PACE incorporates the strong relational inductive bias through a node-dependency-aware positional encoding framework, dag2seq, and a masked self-attention mechanism. Experiments demonstrate that PACE not only generates smooth latent (DAG encoding) space beneficial to the downstream prediction and optimization routines, but also significantly boosts the encoding speed.

\nocite{langley00}

\bibliography{reference}
\bibliographystyle{icml2022}

\newpage
\appendix
\onecolumn

\section{Proof of Theorem~\ref{thm:3.1}}
\label{sec:thm:3.1}
Let $G_{1} = (V_{1}, E_{1}, o_{1})$ and $G_{2} = (V_{2}, E_{2}, o_{2})$ be two labelled graphs, then $G_{1}$ and $G_{2}$ are isomorphic (i.e.$G_{1}$ and $G_{2}$ represent the same computation structure.) if and only if their canonical forms are identical, i.e. $C(G_{1}) = C(G_{2})$. Note that this equation means equality between the canonical forms, not isomorphism. Let $C(G_{1}) = (V^{C}_{1}, E^{C}_{1}, o^{C}_{1})$ and $C(G_{2}) = (V^{C}_{2}, E^{C}_{2}, o^{C}_{2})$. As discussed in section \ref{subsec:dag2seq}, there exists bijections $\pi_{1}:V_{1} \to V^{C}_{1}$ and $\pi_{2}:V_{2} \to V^{C}_{2}$, and we use $\pi^{-1}_{1}: V^{C}_{1} \to V_{1}$ and $\pi^{-1}_{2}: V^{C}_{2} \to V_{2}$ to denote their inverse functions. Then, we have $C(G_{1}) = C(G_{2})$ if and only if (1) $o^{C}_{1}(i) = o^{C}_{2}(i)$ for $\forall i$; and (2) $(i,j) \in E^{C}_{1}$ $\Leftrightarrow $ $(i,j) \in E^{C}_{2}$ for $\forall i, j$.


Next, we will prove Theorem \ref{thm:3.1} by equivelantly showing that the sequence $(o(\pi^{-1}(1))$, $p_{\pi^{-1}(1)})$, $(o(\pi^{-1}(2))$, $p_{\pi^{-1}(2)})$, ..., $(o(\pi^{-1}(n))$, $p_{\pi^{-1}(n)})$ can guarantee the distinctness of canonical forms $C(G)$. For the notation convenience, let function $f(\pi(j), \{\pi(i), (i,j) \in E\}) = \textit{Combine}(\pi(j), \textit{Agg}(\{\pi(i), (i,j) \in E)\})$ be the composition of functions Agg and Combine, then it is injective if and only if both Agg and Combine are injective. Furthermore, we use $\textit{Seq}_{1}$ to denote the sequence $(o_{1}(\pi^{-1}_{1}(1)), p_{\pi^{-1}_{1}(1)}), (o_{1}(\pi^{-1}_{1}(2)), p_{\pi^{-1}_{1}(2)}), ..., (o_{1}(\pi^{-1}_{1}(n)), p_{\pi^{-1}_{1}(n)})$, and $\textit{Seq}_{2}$ to denote the sequence $(o_{2}(\pi^{-1}_{2}(1)), p_{\pi^{-1}_{2}(1)}), (o_{2}(\pi^{-1}_{2}(2)), p_{\pi^{-1}_{2}(2)}), ..., (o_{2}(\pi^{-1}_{2}(n)), p_{\pi^{-1}_{2}(n)})$.

So far, we know $C(G_{1}) \ne C(G_{2})$ $\Leftrightarrow $ there (1) exists $i$ such that $o^{C}_{1}(i) \ne o^{C}_{2}(i) $, or (2) exists $i, j$ such that $(i,j) \in E^{C}_{1}$ but $(i,j) \not\in E^{C}_{2}$ (equivalently, $(i,j) \not\in E^{C}_{1}$ but $(i,j) \in E^{C}_{2}$). 

Now, let's prove $C(G_{1}) \ne C(G_{2})$ $\Rightarrow$ $\textit{Seq}_{1} \ne \textit{Seq}_{2}$.
\begin{itemize}
    \item  \textbf{(1) For the first case}, since $\pi_{1}$, $\pi_{2}$ are the bijections that map $G_{1}$ and $G_{2}$ to their canonical forms, then  we have: 
    \begin{align*}
        o_{1}(\pi^{-1}_{1}(i)) &= o^{C}_{1}(\pi_{1}(\pi^{-1}_{1}(i))) \\
        & = o^{C}_{1}(i) \\
        o_{2}(\pi^{-1}_{2}(i)) &= o^{C}_{2}(\pi_{2}(\pi^{-1}_{2}(i))) \\
        & = o^{C}_{2}(i)
    \end{align*}
    Since $o^{C}_{1}(i) \ne o^{C}_{2}(i)$, then we get $o_{1}(\pi^{-1}_{1}(i)) \ne o_{2}(\pi^{-1}_{2}(i))$, indicating that $\textit{Seq}_{1} \ne \textit{Seq}_{2}$.
    \item \textbf{(2) For the second case}, according to the definition of canonical form, we know that $(\pi^{-1}_{1}(i), \pi^{-1}_{1}(j)) \in E_{1}$ $\Leftrightarrow$ $(i,j) \in E^{C}_{1}$ (similarly, $(\pi^{-1}_{2}(i), \pi^{-1}_{2}(j)) \in E_{2}$ $\Leftrightarrow$ $(i,j) \in E^{C}_{2}$). As such, we get:
    \begin{align*}
        p_{\pi^{-1}_{1}(j)} &= f(\pi_{1}(\pi^{-1}_{1}(j)), \{\pi_{1}(\pi^{-1}_{1}(s)), (\pi^{-1}_{1}(s), \pi^{-1}_{1}(j)) \in E_{1}\}) \\
        &= f(j, \{s, (s, j) \in E^{C}_{1}\}) \\
        p_{\pi^{-1}_{2}(j)} &= f(\pi_{2}(\pi^{-1}_{2}(j)), \{\pi_{2}(\pi^{-1}_{2}(s)), (\pi^{-1}_{2}(s), \pi^{-1}_{2}(j)) \in E_{1}\}) \\
        &= f(j, \{s, (s, j) \in E^{C}_{2}\})
    \end{align*}
    Then, since $(i,j) \in E^{C}_{1}$ but $(i,j) \not\in E^{C}_{2}$, we have $\{s, (s,j) \in E^{C}_{1}\} \ne \{s, (s,j) \in E^{C}_{2}\}$. Since function $f$ is injective, then we have $p_{\pi^{-1}_{1}(j)} \ne p_{\pi^{-1}_{2}(j)}$. Hence, $\textit{Seq}_{1} \ne \textit{Seq}_{2}$
\end{itemize}

In the end, let's prove the other direction, i.e. $\textit{Seq}_{1} \ne \textit{Seq}_{2}$ $\Rightarrow$ $C(G_{1}) \ne C(G_{2})$. When $\textit{Seq}_{1} \ne \textit{Seq}_{2}$, there (1) exists $i$ such that $o_{1}(\pi^{-1}_{1}(i)) \ne o_{2}(\pi^{-1}_{2}(i))$, or (2) exists $j$ such that $p_{\pi^{-1}_{1}(j)} \ne p_{\pi^{-1}_{2}(j)}$.
\begin{itemize}
    \item \textbf{(1) For the first case}, according to the previous analysis, we have:
    \begin{align*}
        o^{C}_{1}(i) &= o_{1}(\pi^{-1}_{1}(i)) \\
        o^{C}_{2}(i) &= o_{2}(\pi^{-1}_{2}(i)) 
    \end{align*}
    Hence, we can get $o^{C}_{1}(i) \ne o^{C}_{2}(i) $, which indicates $C(G_{1}) \ne C(G_{2})$.
    \item \textbf{(2) For the second case}, according to the previous analysis, we get: 
    \begin{align*}
        p_{\pi^{-1}_{1}(j)} &=  f(j, \{s, (s,j) \in E^{C}_{1}\}) \\
        p_{\pi^{-1}_{2}(j)} &= f(j, \{s, (s,j) \in E^{C}_{2}\}) 
    \end{align*}
    Since $f$ is injective, $p_{\pi^{-1}_{1}(j)} \ne p_{\pi^{-1}_{2}(j)}$ implies that $\{s, (s,j) \in E^{C}_{1}\} \ne \{s, (s,j) \in E^{C}_{2}\}$.Then, there must exist $i$ such that $(i,j) \in E^{C}_{1}$ but $(i,j) \not\in E^{C}_{2}$ (or $(i,j) \not\in E^{C}_{1}$ but $(i,j) \in E^{C}_{2}$). Henceforth, we get $C(G_{1}) \ne C(G_{2})$.
\end{itemize}

\section{Mask Matrix}
\label{sec:flo}

Here we provide two potential ways to get the mask matrix in PACE. Following the same notation as the main paper, we use $C(G) = (V^{C}, E^C, o^C)$ to denote the canonical form of the input DAG $G$. 

\paragraph{DFS Algorithm} This algorithm takes the canonical form $C(G)$ as input and performs the DFS (depth first search) algorithm on the graph to explore all the nodes of the graph. Before we start the depth first search, we traverse all edges in $E^C$ to find all direct-successors of each node $i$, and then put them in a set $S(i)$ for $i \in V^{C}$. Then, for each node $i$, we perform the DFS to get a dependent set $\textit{D}(i)$, and we have $M_{j,i} = \textit{False}$ if and only if $j \in \textit{D}(i)$. 

\begin{algorithm}
\caption{DFS Algorithm}
\label{alg: bfs}
\begin{algorithmic}
\STATE {\bfseries Input:} $\textit{D}(i)=\{\}$; $\textit{Visited} = [\textit{False} \ \textit{for}\ j \in V^C]$; a source (start) node $i$, $T = [i]$ (T is a stack).
\STATE $\textit{Visited}[i] = \textit{True}$.
\WHILE {$|T| > 0$.}
\STATE $j = T[-1]$.
\STATE delete $j$ from $T$
\FOR {$k$ in $S(j)$}
\IF {$\textit{Visited}[k] = \textit{Flase}$}
\STATE put $k$ in $\textit{D}(i)$
\STATE $\textit{Visited}[k] = \textit{True}$
\STATE put $k$ in $T$
\ENDIF
\ENDFOR
\ENDWHILE
\end{algorithmic}
\end{algorithm}

\textbf{Floyd Algorithm} The Floyd algorithm is originally proposed for finding shortest paths in directed weighted graphs. Here, we initialize the edge weights to be 1, and implement the Floyd algorithm to find the distance $\textit{dist}(i,j)$ (i.e. length of the shortest directed path) between each node pair $i,j$ in $C(G)$. Then we have $M_{i,j} = \textit{False}$ if and only if $\textit{dist}(i,j) \neq \infty$.     

\begin{algorithm}
\caption{Floyd Algorithm}
\label{alg: floyd}
\begin{algorithmic}
\STATE {\bfseries Input:} $\textit{dist}(i,j)=1$ if $(i,j) \in E^C$ else $\infty$
\FOR{i $\in V^C$}
    \FOR{j $\in V^C$}
        \FOR{k $\in V^C$}
            \IF{$dist(j,k) > dist(j,i) + dist(i,k)$.}
            \STATE $dist(j,k) = dist(j,i) + dist(i,k)$.
            \ENDIF
        \ENDFOR
    \ENDFOR
\ENDFOR
\end{algorithmic}
\end{algorithm}

\section{Multi-Head Self-Attention Mechanism}
\label{sec:multi_head}

Here we introduce the multi-head (masked) self-attention mechanism in the Transformer encoder block of PACE. For the notation convenience, we use $H_k$ to denote the output representation of the $k$th Transformer encoder block, and use $H_0$ to denote the input (i.e. the representation of the sequence generated by dag2seq) to the first Transformer encoder block. Furthermore, we denote the number of heads in the self-attention mechanism as $h$, and the embedding dimension (of each item in the sequence) as $d$. Then the Transformer encoder blocks update representation $H_k$ as following.
\begin{align}
    H^j_k &= \textit{softmax} (\frac{Q^j_k (K^j_k)^T}{d}) V^j_k \ \ \ \  \textit{for} \ j = 1,2, ... h\\
    H_{k+1} &= \textit{feed-forward} (\rVert^{h}_{j=1} H^j_k) 
\end{align}
where $Q^j_k = H_k W^j_{k,q}$, $K^j_k = H_k W^j_{k,k}$, $V^j_k = H_k W^j_{k,v}$ are the query matrix, key matrix and value matrix, respectively (i.e. $W^j_{k,q}, W^j_{k,k}, W^j_{k,v}$ are trainable parameter matrices); $\rVert$ represents the concatenation operation; Feed-forward is a one-layer MLP. When we introduce the mask operation into the Transformer encoder block, let $M$ be the mask matrix from the Floyd algorithm or the BFS algorithm, then we use following equation to replace equation 5 in the Transformer encoder block.
\begin{align}
     H^j_k &= \textit{softmax} (\frac{Q^j_k (K^j_k)^T + -\infty * M}{d}) V^j_k \ \ \ \  \textit{for} \ j = 1,2, ... h
\end{align}



\section{Details about PACE in the VAE Architecture}
\label{sec:overall_detail}

\begin{figure*}[h]
    \centering
    \includegraphics[width=0.9\linewidth]{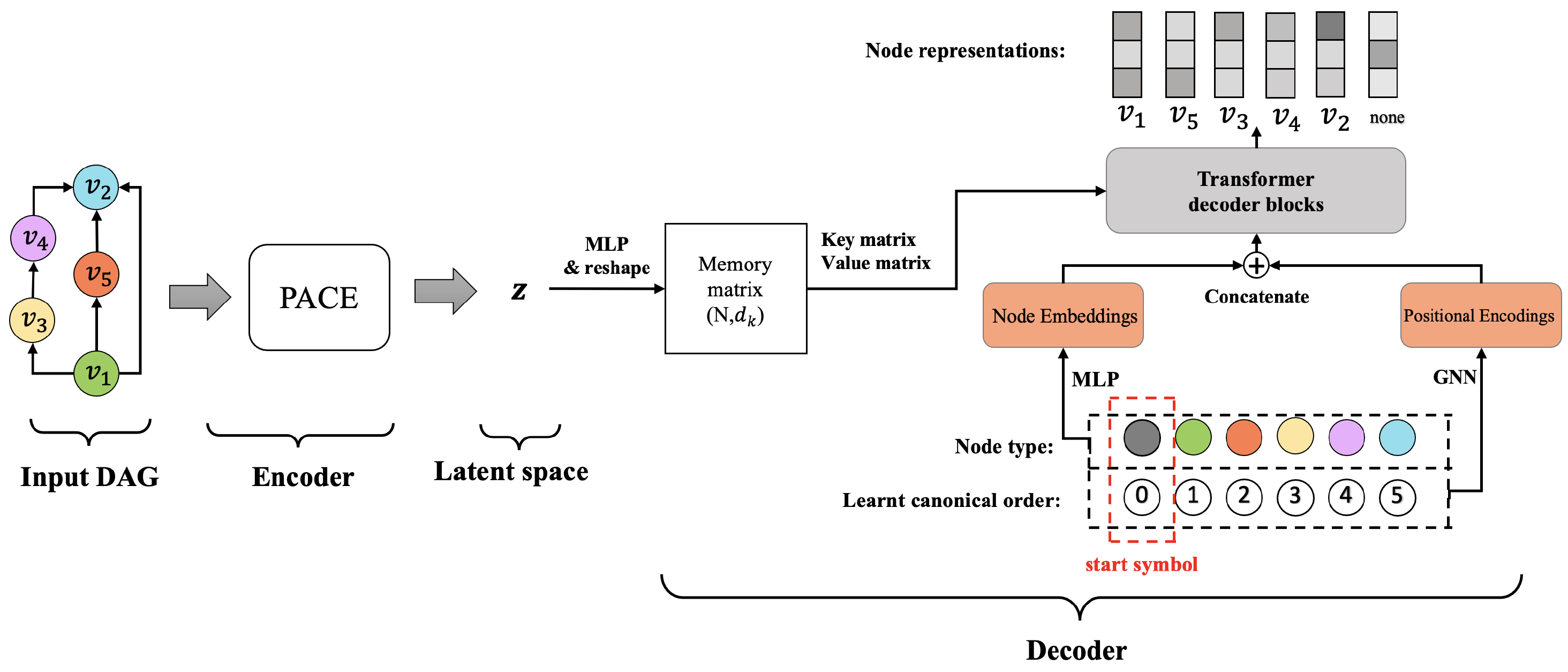}
    \caption{The illustration of PACE in the VAE architecture (PACE-VAE)}
    \label{fig:app fig2}
\end{figure*}

In the section, we describe the decoder of PACE-VAE. Figure \ref{fig:app fig2} illustrates the overall architecture. In the main paper, we have introduced how PACE maps input DAGs to the latent space, here we focus on the decoder of PACE-VAE.

Similar to PACE, the decoder is constructed upon the Transformer decoder block. Each Transformer decoder block consists of a masked multi-head self-attention layer(i.e. Equation 6), a multi-head attention layer (i.e. Equation 4 except that the key matrix and value matrix are computed from points $z$ in the latent space), and a feed-forward layer (i.e. Equation 5).
The decoder takes a MLP as the embedding layer to generate node type embeddings as PACE. In analogous to the dag2seq framework in PACE, the decoder also uses a GNN to generate the positional encoding based on the learnt canonical order of nodes. The node embeddings and positional encodings are concatenated and then fed into multiple stacked Transformer decoder blocks to predict the node representations, which is used to predict the node types and the existence of edges. In analogous to the standard Transformer decoder, the decoder performs the shift right trick (i.e. the $i$th output node representation corresponds to the $i+1$th node in the sequence) and adds a start symbol node (i.e. the black node in Figure \ref{fig:app fig2}) at the beginning of the node sequence. Specifically, the canonical label of the start symbol node is different from any possible canonical label in the dag2seq framework to distinguish it's position. For instance, DAG in the searching space contains at most $N$ nodes, then the canonical order of the start symbol node can be $0$ or $N+1$. Let $o_{i}$ denotes the output representation of node $i$ in the sequence, then it is used to predict the type of node $i+1$ in the sequence through a MLP. Similarly, for any $j < i$, we use another MLP, which takes the concatenation of $o_{j}$ and $o_{i}$ as input, to predict the existence of an directed edge from node $j+1$ to node $i+1$ in the sequence. Note that the canonical order can be generated from the topological sort by breaking ties using canonicalization tools, such as Nauty. Thus, for each node $i$ in the sequence, any dependent node $j$ of this node must be arranged in a prior position in the sequence (i.e. $j < i$). In the end, based on these predictions (node representations), we can perform the teacher forcing to train the VAE.   

Although the ordering of the output sequence can have a significant impact on the performance in the sequence-to-sequence model \cite{vinyals2015order}, it is not the same case for the decoder of PACE-VAE. Let $x_{1}, x_{2}, ...x_{n}$ be an output sequence in the sequence-to-sequence model, and $v_{1}, v_{2}, ...v_{n}$ an output node sequence in the decoder of PACE-VAE. Then, for any $i, j$ such that $i < j$, the sequence-to-sequence model knows that $x_{j}$ is arranged in a later position in the output sequence than $x_{i}$, in other words, $x_{j}$ is dependent on $x_{i}$. However, in the decoder of PACE-VAE, the existence of an edge between $v_{i}$ and $v_{j}$ is predicted by the decoder itself, hence, any topological order is suitable in the decoder of PACE-VAE, and we select (topological) canonical order to facilitate the teacher forcing.

\section{Downstream Search Performance on NA and BN}
\label{sec:visual_detect}

\begin{table*}[h]

\caption{Downstream search performance on NA and BN.}
\label{tab:bo}
\centering
\begin{tabular}{lcccc}
\toprule
Model  & PACE & DAGNN & D-VAE & S-VAE \\
\midrule
(NA) Test accuracy $\uparrow$ & \textbf{95.08} & 95.06  & 94.80 & 92.79 \\
(BN) BIC score $\uparrow$ & \textbf{-11107.29} & \textbf{-11107.29}  & $-$11125.75 & $-$11125.77 \\
\bottomrule
\end{tabular}
\vskip -0.1in
\end{table*}

\begin{figure*}[h]
    \centering
    \includegraphics[width=0.25\linewidth]{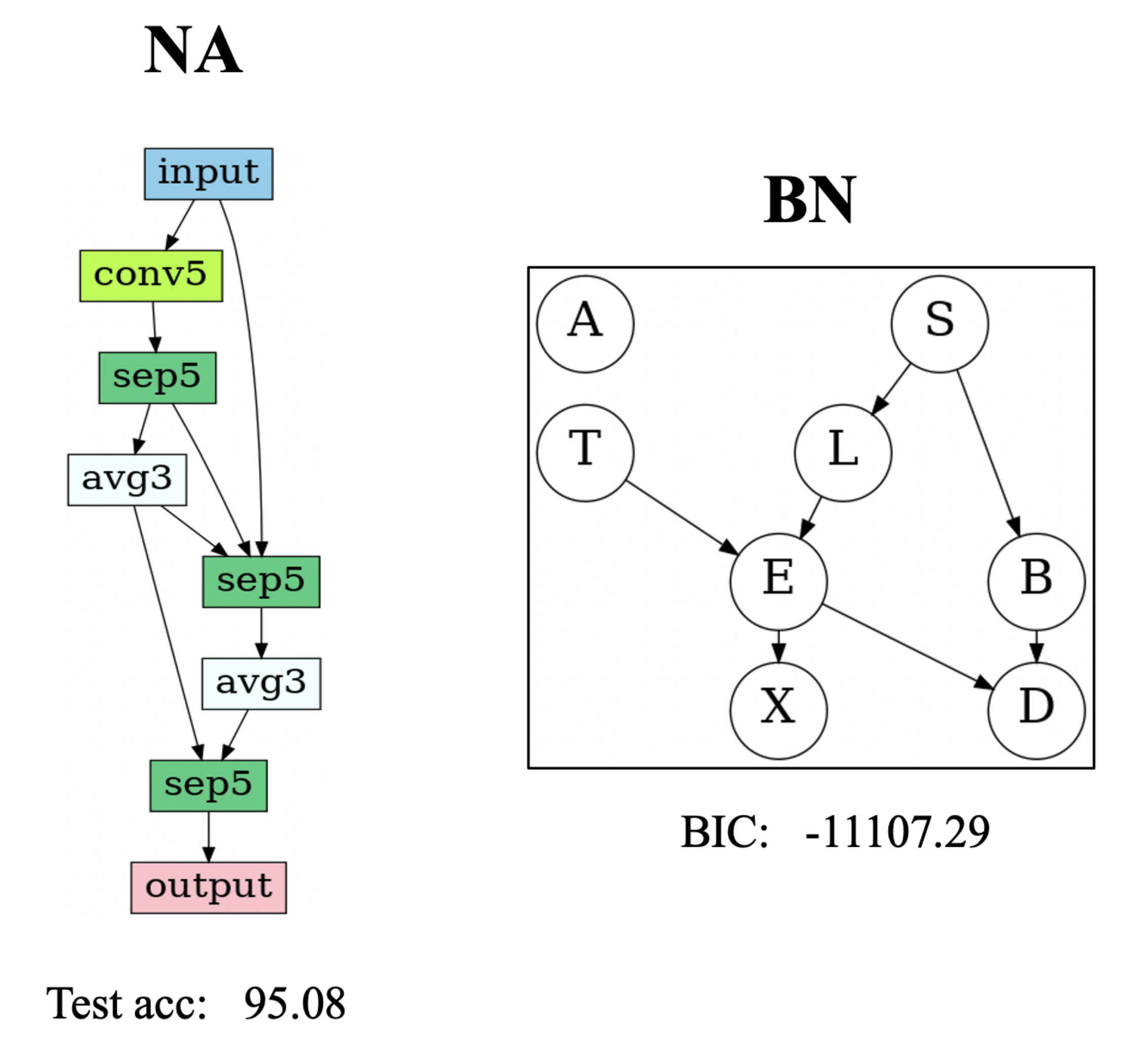}
     \caption{Best architectures on NA and BN detected by PACE.}
    \label{fig: best_arch}
\end{figure*}
In this section, we compare the downstream search performance on datasets NA and BN. As Table \ref{tab:bo} shows, PACE detects architectures with the best performance on both NA and BN. Compared with GRU-based DAG encoders, the Transformer-style attention in PACE has superior ability for learning the long-term dependency. Hence, PACE's latent embeddings are more informative about the computation structure, which favors using a predictive model in its latent space to guide the downstream search routines. 

We also visualize the optimal architectures detected by Bayesian optimization (over the latent DAG encoding space generated by PACE) on datasets NA and BN. Figure \ref{fig: best_arch} illustrates our results. On dataset BN, we find that the detected optimal Bayesian network structure is almost the same as the ground truth (Figure 2 in \citep{lauritzen1988local}), except that there is another directed edge from node A (visit to Asia ?) to node T (Tuberculosis) in the ground truth.



\section{Experiment Details and Further Discussion on OGBG-CODE2}
\label{sec:ogbg-code}
Our paper mainly focuses on the DAG optimization problem. Since searching the optimal neural architectures and Bayesian network structures is essentially typical DAG optimization task, we choose datasets NA, NAS101, NAS301 (neural architecture search benchmark) and BN (Bayesian network structure learning benchmark) for evaluation. However, encoding ASTs (Abstract Syntax Trees) can be another standard application area for DAG-based processing, then we implement additional experiments on dataset OGBG-CODE2 to evaluate PACE against GRU-based DAG encoders. On average, DAGs in dataset OGBG-CODE2 contain more than 120 nodes, while the largest DAG contains more than 30,000 nodes, and the depth of tree can reach 275. Hence, sequentially encoding DAGs in OGBG-CODE2 can be extremely costly, and PACE can take a relative larger number of (self-) attention layers (6 layers) in this setting. On dataset OGBG-CODE2, it's conventionally to use a AST node encoder that incorporates edge features to generate continuous node embeddings. Similarly, PACE uses a GNN as AST node encoder to generate continuous node embeddings, and then dag2seq in PACE approximately takes the DFS order as inputs. 

\textbf{Training method:} As discussed in the main paper, PACE takes the learnt representation of the output node as the graph embedding. On dataset OGBG-CODE2, PACE is not trained in the VAE version nor in the BERT version. Instead, PACE is directly trained on the TOK (token prediction task) in a supervised version based on the graph embedding. Thus, all methods are trained with a single objective and then evaluated on that objective, so that the test F1 score can determine how easily the models can learn. 

In general cases, PACE takes Floyd algorithm \ref{alg: floyd} (see Appendix \ref{sec:flo}) to get the attention mask. However, Floyd algorithm has a time complexity of $O(N^{3})$, where $N$ is the number of nodes in a DAG. Then, the data preprocessing might be time-consuming when applied to large-scale graphs. Hence, on dataset OGBG-CODE2, we propose a backtracking algorithm based on the tree structure of ASTs. The backtracking algorithm has a time complexity of $O(N)$ when the depths of nodes in ASTs are bounded. Then we can significantly reduce the preprocessing time to obtain the attention mask. In algorithm \ref{alg: back_algo}, we use AST $= (V, E)$ to denote an abstract syntax tree, where nodes in $V$ are sorted according to the DFS order. In all experiments, we include the computation time of finding the canonical form of DAGs in the training and inference time of PACE.

\begin{algorithm}
\caption{Backtracking Algorithm}
\label{alg: back_algo}
\begin{algorithmic}
\STATE {\bfseries Input:} backtracking dictionary $\textit{B}=\{\}$, AST $=(V,E)$, function $f: V \to \mathbb{N}$ that maps nodes to the depth, Attention mask $M$ whose elements are set to be True. 
\FOR{$v \in V$}
\STATE $j = f(v)$
\STATE $\textit{B}(j) = v$
\IF{$j > 0$.}
\FOR{$k$ in $1$ to $j$} 
\STATE $M_{\textit{B}(j),\textit{B}(k)} =$ False 
\ENDFOR
\ENDIF
\ENDFOR
\end{algorithmic}
\end{algorithm}



\section{Reconstruction Accuracy and Generation Performance}
\label{sec:recon_and_gene}

\begin{table*}[h]
\caption{Recon. accuracy, valid prior, uniqueness, novelty and overall (ave) performance $\%$}
\label{tab: recon}
\begin{center}
\resizebox{\textwidth}{!}{
\begin{tabular}{lccccccccccccc}
\toprule
& \multicolumn{5}{c}{NA} & & \multicolumn{5}{c}{BN} \\
\cmidrule(r){2-6}  \cmidrule(r){8-12} 
Methods  & Accuracy $\uparrow$ & Valid $\uparrow$ & Unique $\uparrow$ & Novel $\uparrow$ & \textbf{Overall} $\uparrow$ & &Accuracy $\uparrow$ & Valid $\uparrow$ & Unique $\uparrow$ & Novel $\uparrow$  & \textbf{Overall} $\uparrow$\\
\midrule
PACE & 99.97 & 98.16  & 57.77 & 100.00 & \textbf{88.98} & & 99.99 & 99.96  & 45.10 & 98.50 & \textbf{85.88} \\
\midrule
DAGNN & 99.97 & 99.98  & 37.36 & 100.00 & 84.33 & & 99.96 & 99.89  & 37.61 & 98.16 & 83.91\\
D-VAE & 99.96 & 100.00  & 37.26 & 100.00 & 84.31 & & 99.94 & 98.84  & 38.98 & 98.01 & 83.94  \\
S-VAE & 99.98 & 100.00  & 37.03 & 99.99 & 84.25 & & 99.99 & 100.00  & 35.51 & 99.70 & 83.80 \\
GraphRNN & 99.85 & 99.84  & 29.77 & 100.00 & 82.37 & & 96.71 & 100.00  & 27.30 & 98.57 & 80.65 \\
GCN & 5.42 & 99.37  & 41.48 & 100.00 & 61.57 & & 99.07 & 99.89  & 30.53 & 98.26 & 81.94\\
\bottomrule
\end{tabular}
}
\end{center}
\end{table*}
Models parameterized with neural networks contribute to the inductive biases of the deep generative models~\citep{zhang2016understanding,keskar2016large}. Thus, the quality of DAG encoders can also be characterized by the reconstruction accuracy (Accuracy) and the generation performance. (i.e. the proportions of valid/ unique/ novel architectures in generated DAGs) 

The reconstruction accuracy, prior validity, uniqueness and novelty are calculated in the same way as \cite{zhang2019d}. Empirical results are presented in Table \ref{tab: recon}, and we take the average of these four metrics to characterize the overall performance of the deep generative model (i.e. VAE), which also measures the quality of the DAG encoder. We find that PACE performs similarly well in reconstruction accuracy, prior validity and novelty with D-VAE, DAGNN and S-VAE, while significantly improving the uniqueness. Hence, PACE achieves the best overall performance and generates more diverse DAG architectures.


\section{Extension of Theorem \ref{thm:3.1}}

The expressive power of message passing Graph Neural Networks (GNNs) is upper bounded by Weisfeiler-Lehman (1-WL) isomorphism test. To address this limitation, subgraph-based techniques \cite{zhang2021nested,you2021identity} are applied to general message passing GNNs to extend their expressiveness beyond the 1-WL test by learning the representation of each node based on the rooted subgraph around it. Thanks to the plug-and-play framework of these subgraph-based techniques, they share advantages of simplicity and efficiency in various graph learning tasks.

Following this idea, the proposed dag2seq framework can also be extended to a subgraph-based version. Given a DAG $G = (V,E)$ and a node $i$, we use $G^{h}_{i}$ to denote the height-h rooted subgraph of node $i$, which is the subgraph induced from $G$ by the nodes within $h$ hops of node $i$. Let $f$ be a function that maps subgraphs to vectores, then we have the following corollary:

\begin{coro}
\label{coro:3.2}
Let $G = (V,E)$ be a DAG, and $p_{1}, p_{2}, ..., p_{n}$ be the  positional encodings generated by dag2seq. If functions f, \textit{Agg} and \textit{Combine} are injective, then the sequence $(f(G^{h}_{\pi^{-1}(1)})$, $p_{\pi^{-1}(1)})$, $(f(G^{h}_{\pi^{-1}(2)})$, $p_{\pi^{-1}(2)})$, ..., $(f(G^{h}_{\pi^{-1}(n)})$, $p_{\pi^{-1}(n)})$ can injectively represent DAG $G$.  
\end{coro}

As the function $f$ is injective, we can prove corollary \ref{coro:3.2} by simply replacing $o(\pi^{-1}(i))$ in the proof \ref{sec:thm:3.1} by $f(G^{h}_{\pi^{-1}(i)})$ for $\forall i$. Corollary \ref{coro:3.2} shows that the proposed dag2seq framework can be equipped with advancements in GNN literature. When $G$ has a tree structure (e.g. OGBG-CODE), dag2seq can simply take message passing GNN as the function $f$ to injectively represent DAGs as sequences. On the other hand, for general DAGs, the corollary indicates that the more expressive GNNs, such as IDGNN \cite{you2021identity} and NGNN \cite{zhang2021nested}, will generate sequences with higher quality.


\end{document}